\documentclass[runningheads]{llncs}
\usepackage[T1]{fontenc}
\usepackage{hyperref}

\usepackage{graphicx}
\usepackage{color}

\usepackage{booktabs}
\usepackage{graphicx}
\usepackage{amsmath}
\usepackage{amssymb}
\usepackage{microtype}
\usepackage[ruled,vlined]{algorithm2e}
\definecolor{commentcolor}{RGB}{110,154,155}   % define comment color
\newcommand{\PyComment}[1]{\ttfamily\textcolor{commentcolor}{\# #1}}  % add a "#" before the input text "#1"
\newcommand{\PyCode}[1]{\ttfamily\textcolor{black}{#1}} % \ttfamily is the code font
\begin{document}
\title{A Manifold Representation of the Key in Vision Transformers}
%
%\titlerunning{Abbreviated paper title}
% If the paper title is too long for the running head, you can set
% an abbreviated paper title here
%
\author{Li Meng\inst{1}\orcidID{0000-0002-8867-9104} \and
Morten Goodwin\inst{2}\orcidID{0000-0001-6331-702X} \and \\
Anis Yazidi\inst{3}\orcidID{0000-0001-7591-1659} \and
Paal Engelstad\inst{1,3}\orcidID{0009-0000-8371-927X}}
% index{Meng, Li}
% index{Goodwin, Morten}
% index{Yazidi, Anis}
% index{Engelstad, Paal}

\authorrunning{L. Meng et al.}
% First names are abbreviated in the running head.
% If there are more than two authors, 'et al.' is used.
%
\institute{Department of Technology Systems, University of Oslo, Norway \and
Centre for Artificial Intelligence Research, University of Agder, Norway \and
Oslo Metropolitan University, Norway\\
\email{li.meng@its.uio.no}}
\maketitle              % typeset the header of the contribution
\begin{abstract}
Vision Transformers implement multi-head self-attention via stacking multiple attention blocks. The query, key, and value are often intertwined and generated within those blocks via a single, shared linear transformation. This paper explores the concept of disentangling the key from the query and value, and adopting a manifold representation for the key. Our experiments reveal that decoupling and endowing the key with a manifold structure can enhance the model's performance. Specifically, ViT-B exhibits a 0.87\% increase in top-1 accuracy, while Swin-T sees a boost of 0.52\% in top-1 accuracy on the ImageNet-1K dataset, with eight charts in the manifold key. Our approach also yields positive results in object detection and instance segmentation tasks on the COCO dataset. We establish that these performance gains are not merely due to the simplicity of adding more parameters and computations. Future research may investigate strategies for cutting the budget of such representations and aim for further performance improvements based on our findings.

\end{abstract}

\section{Introduction}
In the field of artificial intelligence (AI), the emergence of Transformers \cite{vaswani2017attention} marked a shift in the natural language processing (NLP) landscape, which significantly outpace earlier models such as long short-term memory (LSTM) \cite{hochreiter1997long} and recurrent neural networks (RNNs) \cite{rumelhart1985learning}, \cite{jordan1997serial}. Recently, Vision Transformers (ViTs) \cite{dosovitskiy2020image} have adapted transformers from NLP to computer vision (CV).

While convolution neural networks (CNNs) have played a substantial role in deep learning (DL) \cite{lecun2015deep}, their capability to interpret complex visual scenes is inherently determined by the depth of the model architectures. They rely on predefined convolution and pooling layers to process visual inputs through local, grid-like filters, which can have difficulties when the depth is too shallow to capture meaningful long-range dependencies in the global context. This observation has sparked a growing interest in developing DL models that can more effectively capture global relationships within the visual representations.

In response to this, ViTs propose an alternative that enables the modeling of global contextual information and intricate interdependencies by taking the transformer architecture and applying it to visual images. Unlike CNNs, which process the image inputs through a relatively small kernel, ViTs apply the concept of self-attention, thereby considering the entire image in a holistic view, leading to significant advantages in image understanding.

Subsequent vision transformer models, building on the ViT framework (\cite{han2021transformer}, \cite{pmlr-v139-touvron21a}, \cite{Touvron_2021_ICCV}, \cite{Touvron2022DeiTIR}, \cite{liu2021swin}, \cite{liu2021swinv2}) have achieved state-of-the-art performance in CV. They have started to challenge the long-established supremacy of CNNs and have shown promising capabilities in a wide variety of CV tasks, e.g., image classification and object detection.

\begin{figure}[ht]
\vskip 0.2in
\begin{center}
\centerline{\includegraphics[width=0.8\columnwidth]{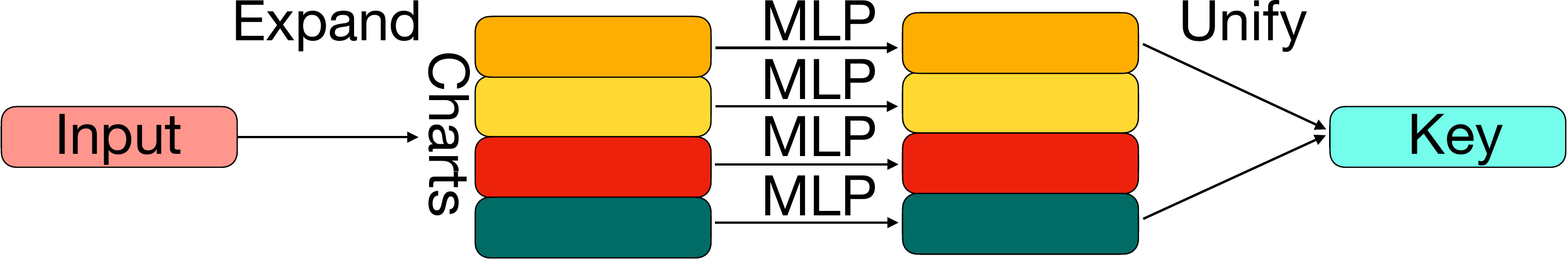}}
\caption{Overview of the proposed architecture. Instead of using a linear layer to generate the key from the input, we expand the input into multiple charts and process these through multilayer perceptrons (MLPs) or MLP-like structures. These representations are then fused to form a meta-representation that serves as the key.}
\label{fig:archi}
\end{center}
\vskip -0.2in
\end{figure}

In prior research, the weights attributed to keys have been handled in a manner analogous to that of values. However, we posit that keys and values are distinct entities and thus merit differential treatment. Investigating the unique mechanisms inherent in keys could pave the way for a more nuanced understanding of vision transformers.

This paper presents a novel and efficacious alteration to the conventional representation of the key in self-attention modules. Our approach is inspired by the unbalanced atlas (UA) paradigm (\cite{korman2021self}, \cite{meng2023state}) from the area of self-supervised learning (SSL). In our methodology, the key's weights have been uncoupled from the weights of the query and value, which are instead characterized as the charts of a manifold. An illustration of our architecture is shown in Fig. \ref{fig:archi}.

In summary, the primary contributions of our work are as follows:
\begin{itemize}
    \item We propose different methods for key disentanglement, which are called SpatialK, KUA, SimpleK, and VanillaK, respectively. Those inventive modifications can improve the performance of transformer models, albeit with increased overhead, and hence deepens our understanding of the key in vision transformers.
    
    \item Through comparative analysis of our proposed key disentanglement methods against baseline models, we demonstrate that enhanced performance is not merely a byproduct of increased computational complexity. In fact, we observe that inappropriate expansion of the key leads to deteriorated model performance. Moreover, our methods are versatile enough to be integrated across various vision transformer frameworks and could potentially serve as a generic module in the future development of efficient, larger-scale transformers.
\end{itemize}

\section{Background}

In the self-attention module of transformer models \cite{vaswani2017attention}, the terms 'query' ($Q$), 'key' ($K$) and 'value' ($V$) are integral components which enable the model to discern and prioritize the significance of different elements within the input tokens, calculated as in Eq. (\ref{eq:attention}), where $d_k$ is the query/key dimension. Keys are used to calculate the attention score in relation to queries. In essence, each element of the input (e.g., word, pixel) is bound to a key, and each key is correlated with a query to yield an attention score. Values, meanwhile, embody the retrieved content decided by the attention scores.
\begin{equation}
    \mathrm{Attention}(Q,K,V) = \mathrm{SoftMax}(QK^T/\sqrt{d_k})V
    \label{eq:attention}
\end{equation}
While multiple keys can assign high attention weights to the same query, it is crucial that their corresponding values remain distinct. For instance, 'social' and 'neural' as keys may both ascribe high weights to the query 'network', but the values for 'social' and 'neural' ought to differ. This underscores our hypothesis that keys function akin to meta-labels, and as such, they stand to gain from a more multifaceted and abstract manner of representation.

\section{Related Work}
In this section, we present a collection of related works that have attempted to refine either the tokens or the self-attention mechanisms, aiming for a more profound understanding and enhanced performance of vision transformers.

\paragraph{Tokens} Marin et al. \cite{marin2023token} discovered that using token pooling to downsample token sets can effectively enhance the computation-accuracy trade-off in the model's performance. The ContextPool method employs adaptive weighting to pool neighboring features for each token before computing attention \cite{huang2022efficient}. Guo et al. \cite{guo2023robustifying} have developed a strategy that integrates average pooling for each token with an attention diversification loss to bolster model robustness. The PatchMerger method merges the number of patches or tokens between two consecutive intermediate layers to diminish the computational demands \cite{renggli2022learning}.

\paragraph{Self-attention} Numerous studies have concentrated on crafting increasingly sophisticated self-attention modules. Retentive Self-Attention (ReSA) infuses distance-related prior knowledge within the architecture through a bidirectional, two-dimensional form of explicit decay \cite{fan2023rmt}. To curb computational complexity, the concept of linear attention has been introduced (\cite{katharopoulos2020transformers}, \cite{wang2020linformer}, \cite{han2023flatten}). The refiner scheme improves the diversity of the self-attention maps through attention expansion and distributed local attention \cite{zhou2021refiner}. Deformable attention advances the learning of sparse-attention patterns, in conjunction with geometric transformations, to enhance the model's performance \cite{xia2022vision}.

\paragraph{Queries} The Query and attend (QnA) technique aggregates the input locally and integrates learned queries, leading to rapid and efficient implementations \cite{arar2022learned}.

\paragraph{Keys} The K-NN attention (KVT) approach distills noise from input tokens by calculating the attention map exclusively with the top-k most similar tokens from the keys for each query \cite{wang2022kvt}. Designating keys as units for dropout acts as a regularization method while also facilitating the probabilistic features of attention weights \cite{li2023dropkey}. Employing a mixture of Gaussian keys within individual attention heads allows for an effective concentration on different parts of the input sequence \cite{nguyen2022improving}.

\paragraph{Manifolds} Multi-manifold multi-head attention characterizes the input space across three unique manifolds, i.e., Euclidean, Symmetric Positive Definite and Grassmann. This approach allows for a more descriptive attention map \cite{konstantinidis2023multi}. Hao et al. \cite{hao2022learning} introduce a manifold distillation methodology and train a compact student model to match a pre-trained teacher model within the patch-level manifold space, equipped with a manifold matching loss.

\section{Method}
A manifold refers to a topological space wherein each point possesses a neighborhood that aligns to local Euclidean properties. Charts play a crucial role in mapping subsets of a manifold to corresponding subsets of Euclidean space. An atlas, then, is a collection of those charts that cover the entire manifold.

Our methodology centers around building a manifold representation of the key in ViTs. First, we broaden the key's dimensionality by inserting a new coordinate dictated by the quantity of charts and introduce diversity among those charts through a trainable parameter $\Gamma$ used for the element-wise multiplication, in a way similar to LayerScale \cite{touvron2021going}. The dimension expansion of input $X$ is described in Eq. (\ref{eq:exp}), where $\odot$ denotes the element-wise product, $H$ the number of charts, $N$ the number of patches, $D$ the hidden dimension, $X\in \mathbb{R}^{N\times D}$, $W\in \mathbb{R}^{(H*D)\times D}$, $\Gamma\in \mathbb{R}^{H*D}$. Afterwards, the expanded key undergoes a group-wise 1D convolution with a kernel size of one.
\begin{equation}
    X = \Gamma\odot XW^T
    \label{eq:exp}
\end{equation}
The employment of group-wise 1D convolutions bears a resemblance to MLPs, as described in \cite{tolstikhin2021mlp}, wherein each chart is treated as a separate group and 1D convolutions are conducted for each of these groups. An output of the 1D convolution with kernel size one can simply be viewed as a linear combination of the input with coefficients along the input channels. The 1D convolution for the $i$-th output channel of $K$ can be described in Eq. (\ref{eq:sp}), where $N$ is the number of input channels, and $\lambda$ is the coefficient.
\begin{equation}
    K_i = \sum_{j=1}^N \lambda_{ij} K_{j}
    \label{eq:sp}
\end{equation}
Moreover, our approach is confined to imitating channel-mixing MLPs to mix features at a given spatial location, omitting the use of token-mixing MLPs. As an alternative, we may adopt the context broadcasting (CB) technique from \cite{hyeon2023scratching}, which serves as a substitute to mix features between different charts.

Ultimately, an aggregation method must be employed to unify the representations from distinct charts into a single meta-representation. One obvious approach to easily combine channel-mixing with aggregation is to forego the use of grouped 1D convolution. Instead, we employ a standard 1D convolution with $H* N$ input channels and $N$ output channels to condense the chart-channel product from $H* N$ to $N$, using Eq. (\ref{eq:sp}). This simplistic approach is denoted as SimpleK.

On the other hand, we explore various aggregation techniques, providing a detailed comparison of their efficacy. Initially, we utilize a linear layer that reshapes the key to match the query's dimension. We refer to this particular approach as SpatialK.

In every instance where the key is computed from input tokens, the conventional linear layer can be substituted with our SpatialK layer. Pytorch-style pseudocode of the SpatialK algorithm is provided in Algorithm \ref{algo:spatialk}.

\begin{algorithm}[ht]
    \PyComment{$x$, $\mathrm{dim}$: input, input dimension} \\
    \PyComment{B, N, H, D: batch, patch, chart and dimension sizes} \\
    \PyComment{$\mathrm{gamma}$: trainable parameter for the element-wise product} \\
    \PyComment{$\mathrm{k}$: $\mathrm{nn.Linear}$(dim, H * D, bias=False)} \\
    \PyComment{$\mathrm{spatial}$: $\mathrm{nn.Conv1d}$(H * N, H * N, kernel\_size=1 groups=H, bias=False)} \\
    \PyComment{$\mathrm{to\_out}$: $\mathrm{nn.Linear}$(H * D, dim, bias=False)} \\
    \PyComment{$\mathrm{rearrange}$: rearrange input to a given shape} \\
    \PyComment{} \\
    \PyComment{key expansion} \\
    \PyCode{$x\; =\;$ gamma * $\mathrm{k}(x)$} \PyComment{B, N, H * D} \\
    \PyCode{$\mathrm{rearrange}$($x$, (B, N * H, D))}\\ 
    \PyComment{1d convolution} \\
    \PyCode{$x\; =\;\mathrm{spatial}(x)$}\\
    \PyCode{$\mathrm{rearrange}$($x$, (B, N, H * D))}\\
    \PyComment{key aggregation} \\
    \PyCode{$x\; =\;\mathrm{to\_out}(x)$}\\

\caption{Pytorch-style pseudocode for SpatialK.}
\label{algo:spatialk}
\end{algorithm}

We combine our SpatialK method with CB to supply the spatial interactions among different charts at a given spatial location, and name this approach as KUA. The calculation for the method with CB is described in Eq. (\ref{eq:cb}), where CB is inserted directly after $\mathrm{spatial}$ in Algorithm \ref{algo:spatialk}. In Eq. (\ref{eq:cb}), $\gamma' \in \mathbb{R}^{D}$ is an additional trainable parameter for the element-wise product.
\begin{equation}
    K = \frac{1}{2}(K + \frac{1}{H*N}\sum_{i=1}^H\sum_{j=1}^N  \gamma'K_{ij})
    \label{eq:cb}
\end{equation}
It is noteworthy that the aggregation of the key does not invariably require a linear output layer. A Minkowski sum of embeddings has been demonstrated as an alternative approach that can competently generate effective aggregated representations \cite{meng2023state}. Building on this insight, we have crafted an additional aggregation strategy that employs a straightforward averaging operation of $K$ with $H$ charts, shown in Eq. (\ref{eq:ua}). This method can serve as a substitute for the 
 $\mathrm{to\_out}$ layer in Algorithm \ref{algo:spatialk}. We integrate the modified approach with CB and name this VanillaK.
\begin{equation}
    K = \frac{1}{H}\sum_{i=1}^H K_{i}
    \label{eq:ua}
\end{equation}
\section{Experimental Details}
We adhere to the customary training protocol from Swin Transformers, aiming to maintain consistency in hyper-parameters across various models to the fullest extent feasible. Our setup employs the AdamW \cite{loshchilov2017decoupled} optimizer with beta values set as (0.9, 0.999), a batch size of 128, and a weight decay factor of 0.05. The learning rate is decayed by a one-cycle cosine scheduling from 5e$^{-4}$ to 5e$^{-6}$ over a span of 300 epochs. For Swin-T, the image and window sizes are configured at 224 and 7, respectively, whereas for SwinV2-T, we adjust these to 256 and 8. For all ViT models in our experiments, the input image size is standardized at 224.

All experiments are conducted using PyTorch \cite{NEURIPS2019_9015} on either four V100 GPUs for the Swin Transformer models or four RTX 2080 Ti GPUs for the ViT models. The ViT models are implemented based on the \textsc{vit-pytorch} library, while the number of FLOPs is counted using the \textsc{fvcore} library.

Given the potential discrepancies in hardware, software, and hyper-parameter configurations compared to the original ViT and Swin Transformer studies, our reported benchmark results may slightly differ. Nevertheless, it is essential to emphasize that those elements were the same for both our proposed methods and our reported benchmarks, ensuring fair and informative comparisons.

\section{Results}
\begin{table*}[t]
\caption{Classification accuracies for ViTs and Swin Transformers on ImageNet-1K. Eight charts are adopted for the manifold key representations.}
\label{tbl:res}
\vskip 0.15in
\begin{center}
\begin{small}
\begin{sc}
\begin{tabular}{lccccr}
\toprule
Method & \# Params [M] & FLOPs [G] & Acc@1 [\%] & Acc@5 [\%] \\
\midrule
ViT-S/16    & 22& 4.7& 78.23 & 93.92\\
+SpatialK      & 52& 11.3&   78.65  & 94.18 \\
+KUA      & 52& 11.3&  78.74 &  94.23  \\
+SimpleK      & 38& 8.6 & 78.34  &   93.98 \\
+VanillaK      & 38& 8.6 & \textbf{78.8}  &   94.1 \\
\midrule
ViT-B/16    & 87& 17.7 & 79.44 & 93.91\\
+SpatialK      & 197& 41.5&  78.69   & 93.93 \\
+KUA      & 197& 41.5 & 78.41  &  93.6  \\
+SimpleK     & 140& 30.4 &    78.68 & 93.87 \\
+VanillaK     & 140& 30.4 &    \textbf{80.31} & 94.54 \\
\midrule
Swin-T    & 28& 4.6& 80.99 & 95.5\\
+SpatialK      & 61& 10.4 &  \textbf{81.51}   & 95.69  \\
+KUA      & 61& 10.4 &  81.26 & 95.58 \\
+SimpleK      & 44& 7.6&  81.31  & 95.55  \\
+VanillaK      & 44& 7.6&  81.29  & 95.53  \\
% \midrule
% SwinV2-T    & 28 & 6.1& 81.94 &95.91\\
% +SpatialK      & 61 & 13.8&  \textbf{81.97}   & 95.86 \\
\bottomrule
\end{tabular}
\end{sc}
\end{small}
\end{center}
\vskip -0.1in
\end{table*}

The full results of ViTs and Swin Transformers on the ImageNet-1K dataset are thoroughly detailed in Table \ref{tbl:res}. The reported top-5 score is from the model that obtains the highest top-1 score over the course of 300 epochs. The results underscore that, out of SpatialK, KUA and VanillaK, the most effective method varies between ViTs and Swin Transformers.

Surprisingly, for ViTs, a simple averaging aggregation scheme of VanillaK proves to be quite successful. VanillaK obtains a top-1 score of 78.8\% when applied to ViT-S models, and 80.31\% when applied to ViT-B models, which are the highest among the presented results of ViT models in Table \ref{tbl:res}. Their top-1 scores are impressively higher than those of the baselines, which are 78.23\% and 78.8\%, yielding 0.57\% and 0.87\% increases when applying VanillaK to ViT-S and ViT-B, respectively.

While SimpleK does not surpass VanillaK in achieving the best top-1 scores due to its simplicity, it nonetheless demonstrates improvement on the ViT-S and Swin-T models, with scores of 78.34\% and 81.31\%, respectively, marking 0.11\% and 0.32\% increases.

In contrast, the SpatialK method shines in the context of Swin Transformers, attaining a top-1 accuracy of 81.51\%, which is 0.52\% higher than the reported Swin-T benchmark results. This indicates that, for Swin Transformers, a linear technique for aggregating manifold representations is beneficial.

\subsection{Charts and CB}
\begin{table}[t]
\caption{Classification accuracies for ViT-B/16 on ImageNet-1K, experimenting on the number of charts, adding VanillaK without CB.}
\label{tbl:vitbk}
\vskip 0.15in
\begin{center}
\begin{small}
\begin{sc}
\begin{tabular}{lcccr}
\toprule
Charts & \# Params [M] & Acc@1 [\%] & Acc@5 [\%] \\
\midrule
4    & 110& \textbf{80.24}&      94.43   \\
8     & 140& 78.69& 93.84\\
\bottomrule
\end{tabular}
\end{sc}
\end{small}
\end{center}
\vskip -0.1in
\end{table}

We examine the impact of varying the number of charts of ViT-B models in Table \ref{tbl:vitbk}. Notably, ViT-B manifests an improved top-1 accuracy of 80.24\% with four charts as opposed to 78.69\% with eight charts. This demonstrates that our model's advancements are not merely attributable to increased complexity. Instead, astutely tailored aggregation methods significantly enhance the model's performance. The presence of CB is vital in allowing the performance of VanillaK to scale in a cost-effective manner with a large number of charts.

\subsection{Visualization}
\begin{figure*}[ht]
\vskip 0.2in
\begin{center}
\centerline{\includegraphics[width=0.8\textwidth]{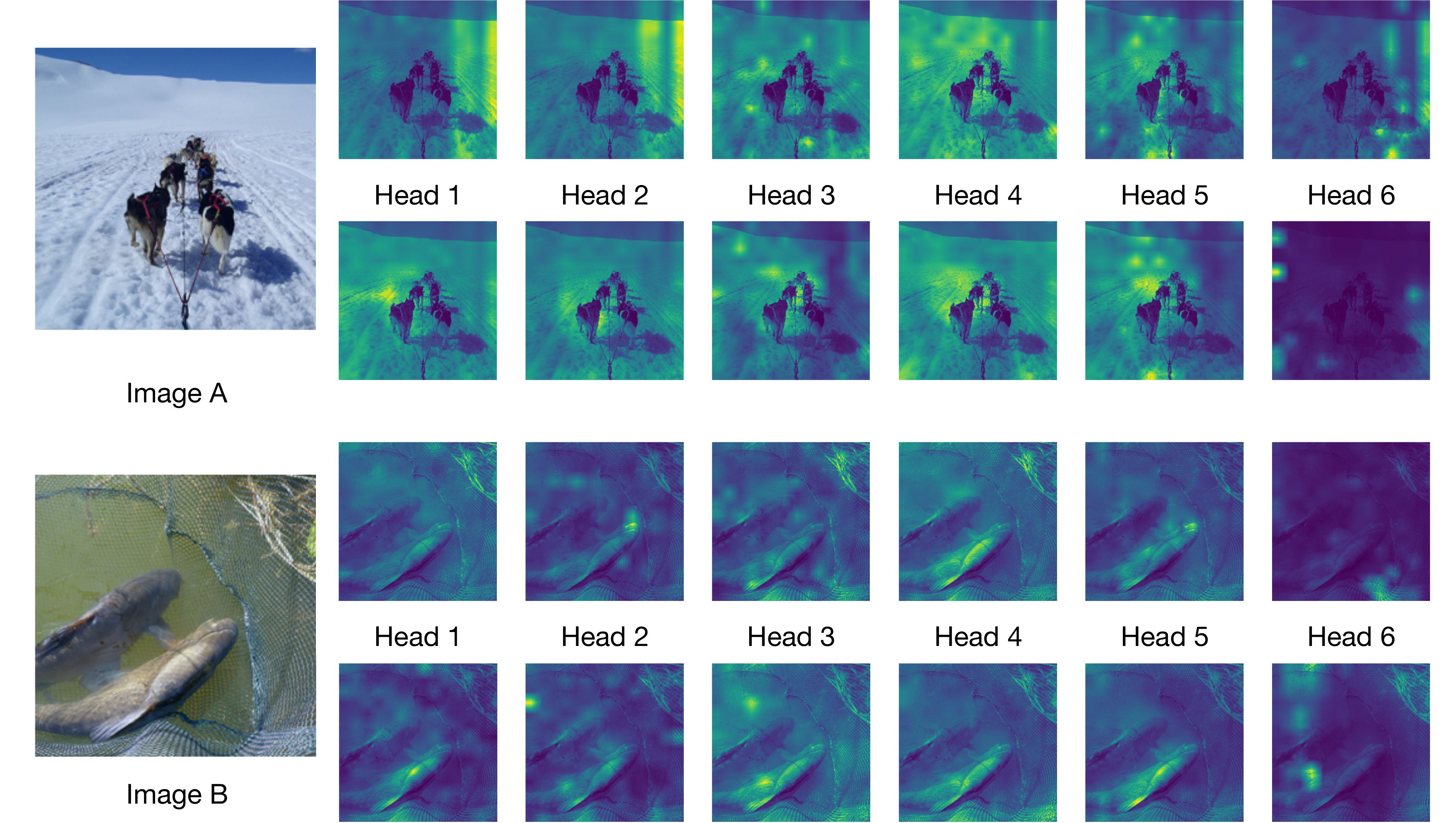}}
\caption{Attention maps of the last layer for each head. We compute the joint attention for each head separately using ViT-S/16 models pre-trained on ImageNet-1K. The first row for each original image contains the attention maps of the baseline ViT-S model, and the second row contains the attention maps of the ViT-S model with VanillaK.}
\label{fig:attn}
\end{center}
\vskip -0.2in
\end{figure*}

Fig. \ref{fig:attn} showcases the activation maps of the joint attention for each head in the final layer. We can observe a more relevant attention positioning towards the image content rather than the background after implementing VanillaK. Taking the instance of image A, the joint attention is primarily concentrated on the snow-covered region on the image's right side without VanillaK, but concentrated on the sleigh instead after adding VanillaK for the first two heads. Moreover, only half of the heads focus attention on the sleigh without introducing a manifold key representation, whereas 5 out of the 6 heads is able to direct attention to the sleigh upon the integration of VanillaK.

This is also the case in image B. When VanillaK is introduced, the attendance to the fish object becomes more salient, particularly in heads 3 to 5. Analogous to the attention diversion to the snow in image A, the focus on the fishnet in the background of image B is also witnessed. However, this background attention is reduced following the implementation of VanillaK.

\subsection{Object Detection}
We implement our object detection and instance segmentation methods using \textsc{MMDetection} \cite{mmdetection}, evaluating on the COCO dataset. Swin-T methods are implemented together with Mask R-CNN \cite{he2017mask} with FPN \cite{lin2017feature}. We fine-tune the pretrained weights for 36 epochs (3$\times$ schedule). The AdamW optimizer has betas of (0.9, 0.999), a batch size of 128, and a weight decay coefficient of 0.05. The learning rate followed a linear decay from 1e$^{-4}$ to 1e$^{-6}$.

Table \ref{tbl:coco} presents the results for both the object detection and instance segmentation tasks. Evidently, the employment of SpatialK, incorporating eight charts, leads to marked enhancements in benchmark performance metrics. For object detection, SpatialK yields improvements of 0.5\%, 0.2\%, and 0.4\% in $\text{AP}^b$, $\text{AP}^b_{50}$, and $\text{AP}^b_{75}$, respectively. Similarly, in the domain of instance segmentation, we observe strides of 0.3\%, 0.5\%, and 0.5\% for $\text{AP}^m$, $\text{AP}^m_{50}$, and $\text{AP}^m_{75}$, respectively. These increments denote significant progress in the performance of object detection and instance segmentation.

\begin{table*}[t]
\caption{COCO object detection and instance segmentation performance on Swin-T models, fine-tuned for 36 epochs (3$\times$ schedule).}
\label{tbl:coco}
\vskip 0.15in
\begin{center}
\begin{small}
\begin{sc}
\begin{tabular}{lccccccr}
\toprule
Method & $\text{AP}^b$ [\%] & $\text{AP}^b_{50}$ [\%]& $\text{AP}^b_{75}$ [\%] & $\text{AP}^m$ [\%]& $\text{AP}^m_{50}$ [\%]& $\text{AP}^m_{75}$ [\%]\\
\midrule
Swin-T    & 45& 67.1& 49.5& 40.7 & 63.8& 43.6   \\
+SpatialK   & \textbf{45.5}& \textbf{67.3}& \textbf{49.9}& \textbf{41} & \textbf{64.3}& \textbf{44.1} \\
\bottomrule
\end{tabular}
\end{sc}
\end{small}
\end{center}
\vskip -0.1in
\end{table*}

\section{Discussion}
Our work has successfully demonstrated the potential for enhanced performance through the development of sophisticated manifold representations of the key within vision transformers. The merits and demerits of our methodologies are distinct. Employing a significantly larger model might not be fully justified in some circumstances. However, the drawbacks associated with increased model complexity can be mitigated, and there are strategies to circumvent the escalation in computational burden. For instance, adopting key representations similar to learned queries \cite{arar2022learned}, as opposed to expanding keys via a linear layer, may result in a substantial reduction of both the count of parameters and the computational cost measured in FLOPs. Meanwhile, a straightforward aggregation technique using a small budget, i.e., VanillaK, have proven to be surprisingly effective.

We have noted that our baseline results fall marginally short of those achieved by the Swin models. This discrepancy primarily stems from issues encountered while attempting to replicate the original work under varied experimental settings. It is essential to highlight that all comparisons between our proposed methods and the baselines were made under identical hardware and software conditions. Therefore, we believe the outcomes are adequate to demonstrate the significant impact of key disentanglement.

By analyzing the intrinsic dimension (ID) in transformers, Valeriani et al. \cite{valeriani2023geometry} discovered that data initially spans a high-dimensional manifold in the earliest layers but then undergoes considerable contraction. These insights pave the way for potentially differentiated strategies in handling manifold representations at various depths of a network. Our proposed methods may also potentially benefit from the Aggregated Pixel-focused Attention \cite{shi2023transnext}, which integrates query embedding and positional attention mechanisms with pixel-centric attention.

In sum, the methodologies we have introduced lay the groundwork for leveraging more intricate and abstract key representations to improve the performance of vision transformers. At present, the computational overhead appears to render the adaptation impractical for some cases. Nonetheless, with the deeper understanding of key disentanglement provided by this paper, there remains an open field for continued exploration, specifically aimed at reducing computational expenses and creating even more potent representations.

\begin{credits}
\subsubsection{\ackname}
This work was performed on the resources from the Centre for Artificial Intelligence Research, University of Agder, and the Department for Research Computing at USIT, University of Oslo.
\end{credits}

\bibliographystyle{splncs04}
\bibliography{main}

%\section*{Appendix}
\end{document}